\documentclass[letterpaper, 10 pt, conference]{ieeeconf}  

\usepackage[utf8]{inputenc}

\IEEEoverridecommandlockouts                              

\usepackage{graphics} 
\usepackage{epsfig} 
\usepackage{mathptmx} 
\usepackage{times} 
\usepackage{amsmath} 
\usepackage{amssymb}  

\usepackage{enumerate}

\usepackage[caption=false]{subfig}

\usepackage{url}
\usepackage[acronym]{glossaries}
\usepackage{multirow}
\usepackage{makecell}
\usepackage{placeins}
\usepackage{xcolor}
\usepackage{supertabular,booktabs}
\usepackage{tabularx}
\usepackage{balance}
\usepackage[colorinlistoftodos,prependcaption,textsize=tiny]{todonotes}
\newcolumntype{Y}{>{\centering\arraybackslash}X}

\usepackage{tikz}
\usepackage{textcomp}
\usepackage{hyperref}
\usepackage{lipsum}

\newcommand\copyrighttext{%
  \footnotesize \textcopyright 2019 IEEE. Personal use of this material is permitted.
  Permission from IEEE must be obtained for all other uses, in any current or future
  media, including reprinting/republishing this material for advertising or promotional
  purposes, creating new collective works, for resale or redistribution to servers or
  lists, or reuse of any copyrighted component of this work in other works.
  }
\newcommand\copyrightnotice{%
\begin{tikzpicture}[remember picture,overlay]
\node[anchor=south,yshift=10pt] at (current page.south) {\fbox{\parbox{\dimexpr\textwidth-\fboxsep-\fboxrule\relax}{\copyrighttext}}};
\end{tikzpicture}%
}

\title{\LARGE \bf
ANDA: A Novel Data Augmentation Technique\\Applied to Salient Object Detection
}

\author{Daniel V. Ruiz, Bruno A. Krinski, and Eduardo Todt \\
   Department of Informatics, Federal University of Paran\'a (UFPR), Curitiba, PR,  Brazil 
   \\ \textit{\{dvruiz, bakrinski, todt\}@inf.ufpr.br} }

\begin{document}

\maketitle
\copyrightnotice

\newacronym{knn}{kNN}{K-Nearest Neighbors}
\newacronym{sod}{SOD}{Salient Object Detection}
\newacronym{gan}{GAN}{Generative Adversarial Network}
\newacronym{lbp}{LBP}{Local Binary Patterns}
\newacronym{fcn}{FCN}{Fully Convolutional Network}
\newacronym{maskrcnn}{Mask-RCNN}{Mask Regional Convolutional Neural Network}
\newacronym{resnet}{ResNet}{Residual Network}
\newacronym{mae}{MAE}{Mean Absolute Error}

\newacronym{svm}{SVM}{Support Vector Machine}
\newacronym{rpn}{RPN}{Region Proposal Network}

\newacronym{cnn}{CNN}{Convolutional Neural Network}
\newacronym{hed}{HED}{Holistically-Nested Edge Detector}
\newacronym{slic}{SLIC}{Simple Linear Iterative Clustering}
\newacronym{roialign}{RoIAlign}{Region of Interest Alignment}
\newacronym{fasterrcnn}{Faster-RCNN}{Faster Regional Convolutional Neural Network}

\thispagestyle{empty}
\pagestyle{empty}

\begin{abstract}

In this paper, we propose a novel data augmentation technique (ANDA) applied to the \gls*{sod} context. Standard data augmentation techniques proposed in the literature, such as image cropping, rotation, flipping, and resizing, only generate variations of the existing examples, providing a limited generalization. Our method has the novelty of creating new images, by combining an object with a new background while retaining part of its salience in this new context; To do so, the ANDA technique relies on the linear combination between labeled salient objects and new backgrounds, generated by removing the original salient object in a process known as image inpainting. Our proposed technique allows for more precise control of the object’s position and size while preserving background information. Aiming to evaluate our proposed method, we trained multiple deep neural networks and compared the effect that our technique has in each one. We also compared our method with other data augmentation techniques. Our findings show that depending on the network improvement can be up to 14.1\% in the F-measure and decay of up to 2.6\% in the  Mean Absolute Error.

\end{abstract}


\glsresetall
\section{Introduction}

Visual Salience (or Visual Saliency) is the characteristic of some objects that makes them stand out from its surrounding regions and attract the attention of the human brain~\cite{Itti1998}. Light intensity,  edge or line orientation,  color,  motion,  and stereo disparity are examples of Visual Salience features that attract human attention.  It has a wide range of applications in  Computer  Vision and  Image  Processing,  e.g.,  recognizing and tracking objects, image cropping and resizing, video and image compression and summarization~\cite{lee:deep_saliency,xi:a_fast}.  In robotics, it can be applied in a plethora of algorithms such as robot self-localization in unknown environments~\cite{todt:outdoor_landmark}, find potential gas/odor sources~\cite{Ishida2006} and as a visual cue for gaze shift as shown by \cite{Orabona2005robs1,Kirchner2011,Shon2005}. 

The recent works in the \gls*{sod} literature proposed the use of Deep Neural Networks to find the salient objects in images~\cite{borji:survey}. Besides the impressive results achieved, training deep learning methods requires vast amounts of data. The construction of large datasets is challenging, especially because of the manual labor demanded. To counter this issue, data augmentation is often applied to multiply the available labeled data. In the \gls*{sod} field, data augmentation techniques are usually limited to image cropping and affine transformations. When relying purely on those operations, background information can be lost, and this lost data could be meaningful in segmentation learning.

In this paper, we propose a novel approach to data augmentation in the context of  \gls{sod}.  Our method uses a linear combination between a labeled salient object and a  new background generated by removing the original salient object. This technique allows more precise control of the object's position and size in the image while preserving the background information. The implementation of the method is publicly available\footnote{https://github.com/ruizvitor/ANDA}.

We present our findings on four distinct neural networks: three \glspl*{fcn} with \gls*{resnet}-50, \gls*{resnet}-101, and VGG-16 backbones, implemented on KittiSeg framework\footnote{https://github.com/MarvinTeichmann/KittiSeg}~\cite{teichmann:kittiseg}, following Krinski~\emph{ et. al}~\cite{krinski2019}; and the PoolNet\footnote{https://github.com/backseason/PoolNet}~\cite{Liu2019PoolSal} with ResNet-50 backbone (a very recent network which achieved an
impressive state of the art results in the \gls*{sod}). The official code released by the authors of PoolNet was used in this work. We also compare our approach to other augmentation techniques proposed in the literature, such as horizontal flipping, rescale, rotation, and random cropping. Eight training sets were produced, as described in Section~\ref{sec:exp}. For each network, cross-dataset tests were performed in eight different publicly available datasets. Each experiment repeated
those tests varying only the training set.

\section{Related Work}
In  this  section,  we  briefly  survey  the  data  augmentation techniques used in recent SOD literature. Perazzi \emph{et al.}~\cite{Perazzi2017CVPR} proposed a data augmentation based on an affine 
transformation of scale and translation. Transformation of scale is applied to generate objects
with $\pm5\%$ of the original size and transformation of translation, shifting the objects with
$\pm10\%$ from the original position. Thin-plate splines are also utilized to generate non-rigid deformations in the width and height of the salient objects. Bianco \emph{et al.}~\cite{Bianco2017RC} proposed a data augmentation based on random crop, where a square subfigure was chosen with a minimum size of 256 and up to the original size then resized to $256\times256$, random horizontal flip, and random change on luminance (gamma). 

Aytekin \emph{et al.}~\cite{Aytekin2018} generate five new images for each image in the training and validation sets with a super-pixel gridization method. Randomly flipping in the horizontal direction is another augmentation technique evaluated. Guo \emph{et al.}~\cite{GuoSymmetry2018} evaluated data augmentation techniques based on random rotation, horizontal and vertical shift, and horizontal and vertical flip on the ICOSEG~\cite{icoseg} dataset. Liu \emph{et al.}~\cite{Liu2019PoolSal} also utilized a random horizontal flip to perform data augmentation. Huang \emph{et al.}~\cite{Huang2019} proposed a data augmentation based on image crop. The bounding boxes of the salient objects are utilized to randomly sample five start and end positions in a way that the cropped images cover the entire salient object. They also applied augmentation techniques like horizontal flip in the cropped images, generating ten new samples per original training image. Similarly, Laroca et al.~\cite{laroca2019} exploited various data augmentation strategies, such as random flipping, rescaling, shearing, and cropping, to train their networks and avoid~overfitting.

\begin{figure*}[t]
\centering

\includegraphics[width=0.99\textwidth]{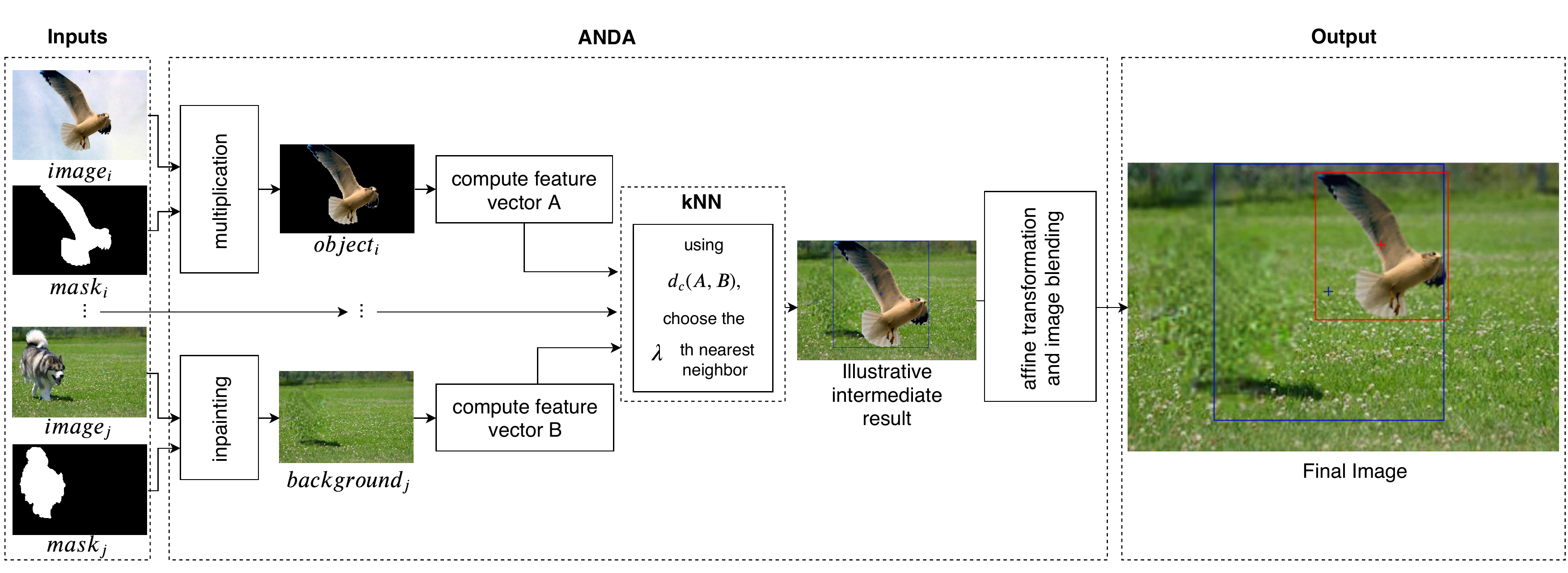}

\caption{Overview diagram of the ANDA technique. The first step is the computation of a background image for all the images of the inputted dataset, an object image is also computed; those images are used to produce feature vectors that can be compared using kNN with cosine similarity; $\lambda$ stands for the criteria chosen. Three different criteria were evaluated, see Section~\ref{sec:exp}. Finally, for each pair of $object_i$ and $background_\lambda$, a final image is created using a linear combination of both. Note that in $mask_j$ the dog's shadow is not included, so $background_j$ preserve this information. In the final image and the intermediate result, blue represents the original bounding box; red represents the bounding box after an affine transformation; the crosses represent the respective center of the bounding boxes.}
\label{fig:fullexample}
\end{figure*}

\section{Proposed Work}\label{sec:proposedWork}

In the first step of the ANDA method, all salient objects are removed from all images in the training
dataset, in turn, generating a new background for each training image (details presented in
Section~\ref{backgroundImage}). The second step is the association of a salient object to a new
background without objects inside (details presented in Section~\ref{linearObjBg}). In order to
preserve the saliency of the object when inserted into the new context, a proper background is chosen
based on the technique described in Section~\ref{knn}.

Changing the background of a salient object is not enough to further the generalization of the training dataset. So, an analysis of the position and size distribution of the labeled objects on multiple datasets used in the SOD literature was performed, see Fig.~\ref{fig:dist}. After analyzing the position distribution, a lack of samples closer to the margins was also observed, see Fig.~\ref{subfig:msra10kpos}. A concentration in the proportional size of the object to the entire image was observed on the MRSA10K dataset~\cite{msra10k}, see Fig.~\ref{subfig:msra10ksize}. Those distributions undermine  its generalization, so to produce new samples that are diverse in size and position, affine transformations were used in the labeled objects before associating objects in a new background. Further details are presented in Section~\ref{size_position}.

\subsection{Background Image Generation}
\label{backgroundImage}

Part of the novelty of this method is the use of a technique called image inpainting, which is the process of restoring missing pixels in digital images in a plausible way. Research in image inpainting has received considerable attention in different areas, such as restoration of old and damaged documents,  removal of unwanted objects, and retouching applications~\cite{QURESHI2017177}. More specifically, we use the neural network architecture named PConv proposed by Liu \emph{et al.}~\cite{nvidiaInpainting}. The PConv is a UNet-like architecture~\cite{unet2015} and was used in our work to remove the labeled salient object of each image of the MSRA10K dataset, creating background only images. In this work, a variation of an unofficial implementation was used with pre-trained weights from ImageNet~\cite{imagenet_cvpr09}, an example of full remotion is shown in Fig.~\ref{fig:fullexample}.

\subsection{Linear Combination of object and background}
\label{linearObjBg}
When combining an object $o$ with and background only image $b$, $o$ and $b$ are padded, ensuring
that both have the same width and height. The ground-truth mask $m_o$ (composed by zeros and ones,
where one is a salient pixel) of the object $o$, is dilated using a $3\times3$ kernel and
mean blurred using a $3\times3$ kernel to achieve a smooth transition between the object and the
background. The $\alpha$ transparency of the object is changed by multiplying $m_o$ by $0.95$. The resulting image $r$ is cropped to $b$ original width and height.

\subsection{Background Choosing}
\label{knn}
When a salient object is inserted in a new background, the salient object may no longer be salient.
In order to choose a background which maintains the object's salience, we compute a feature
vector of 256 positions composed of four histograms (64 bins for Hue, 64 for Saturation, 64 for
Value, and 64 for \gls*{lbp}~\cite{lbp}, with the parameters: number of circularly symmetric neighbour set points: 24, circle radius: 3, and method: uniform) for the salient object and the new background.

A \gls*{knn} with $k=10,000$ ($10,000$ is the number of images in the MSRA10K dataset) is
computed using the cosine similarity defined at Equation~\ref{eq:cosine}. The similarity value is obtained  by using the feature vector $A$ (the salient object without its original background) and the feature vector $B$ (new background image). In this way $d_c(A,B)$ shows the similarity between the object and the background. 

\begin{equation}
    d_c(A,B)=\frac{\sum_{j=1}^{N}A_jB_j}{\sqrt{\sum^{N}_{j=1}A_{j}^{2}}\sqrt{\sum^{N}_{j=1}B_{j}^{2}}}
\label{eq:cosine}
\end{equation}

The $\left \lfloor{ \frac{k}{2} }\right \rfloor$th nearest neighbor was chosen to preserve the saliency of the generated image. This particular value was adopted since the images with higher values are very different, while those with lower values are very similar, in this way a middle ground between those extremes seems to be the best choice, producing images that are closer to the real-world samples. In Section~\ref{sec:exp}, three experiments were performed to evaluate this assumption.

\begin{figure*}[!t]
\centering

\subfloat[]{\includegraphics[width=0.24\textwidth]{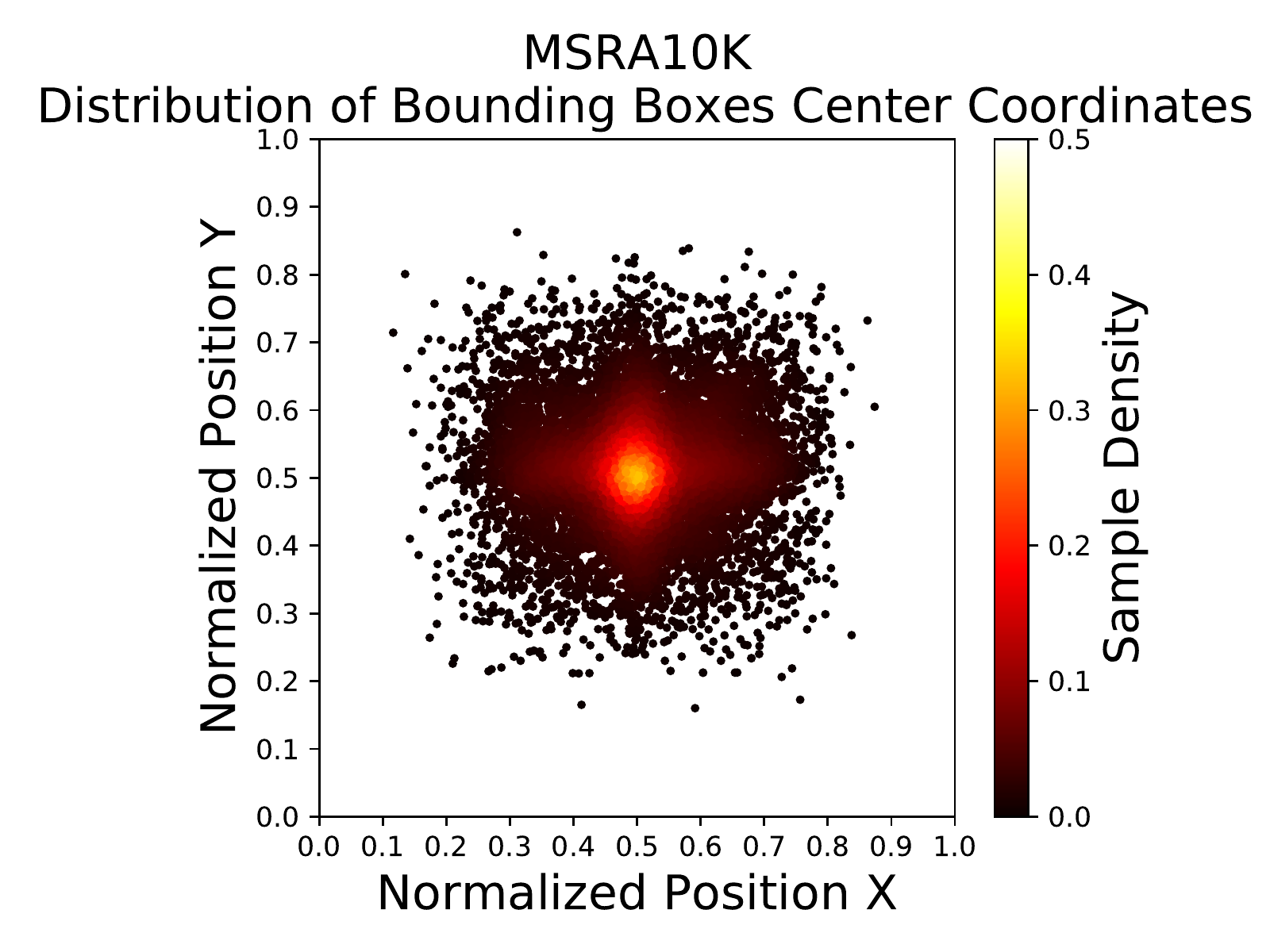} 
\label{subfig:msra10kpos}}
\hfil
\subfloat[]{\includegraphics[width=0.24\textwidth]{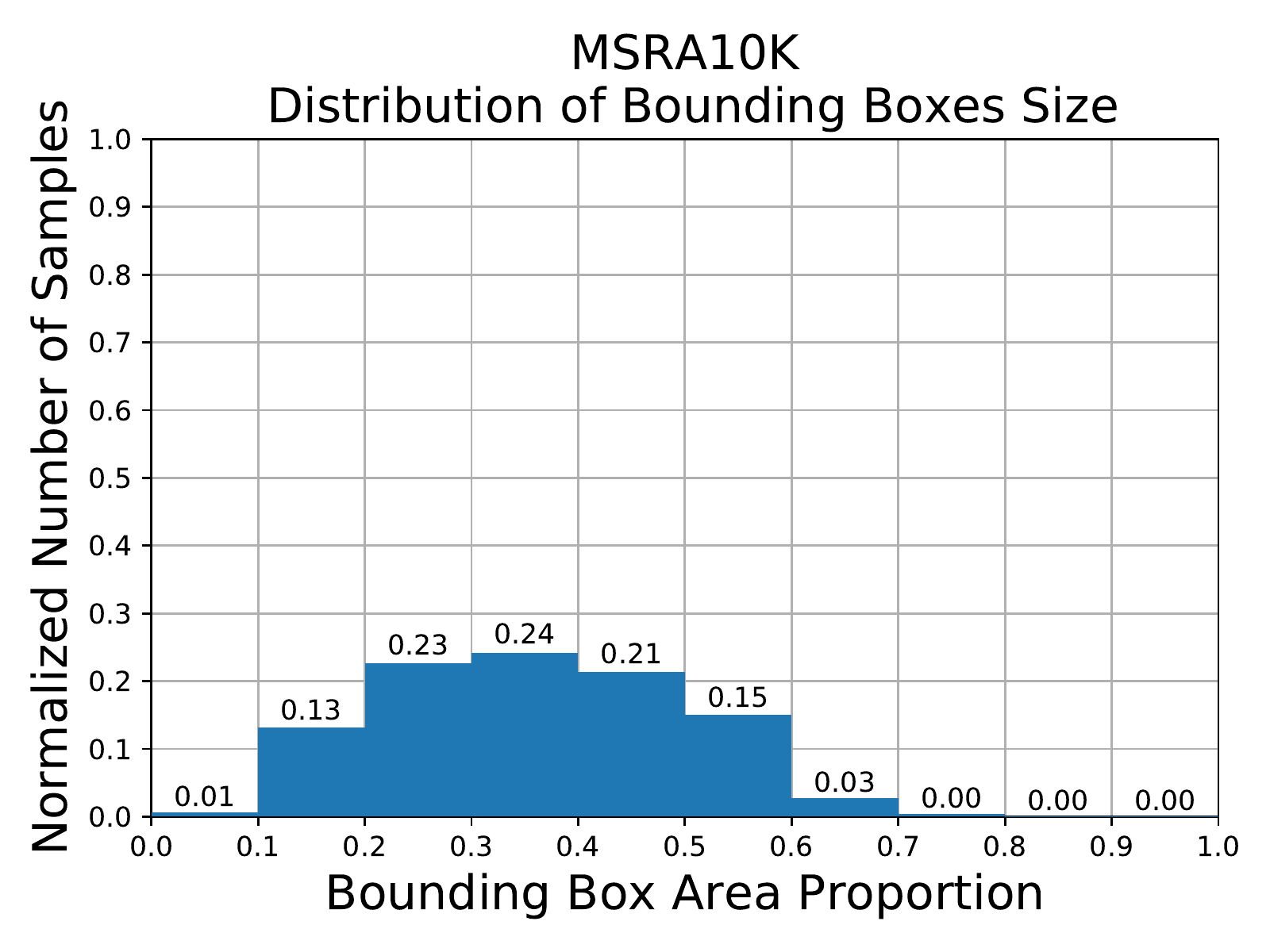}
\label{subfig:msra10ksize}}
\hfil
\subfloat[]{\includegraphics[width=0.24\textwidth]{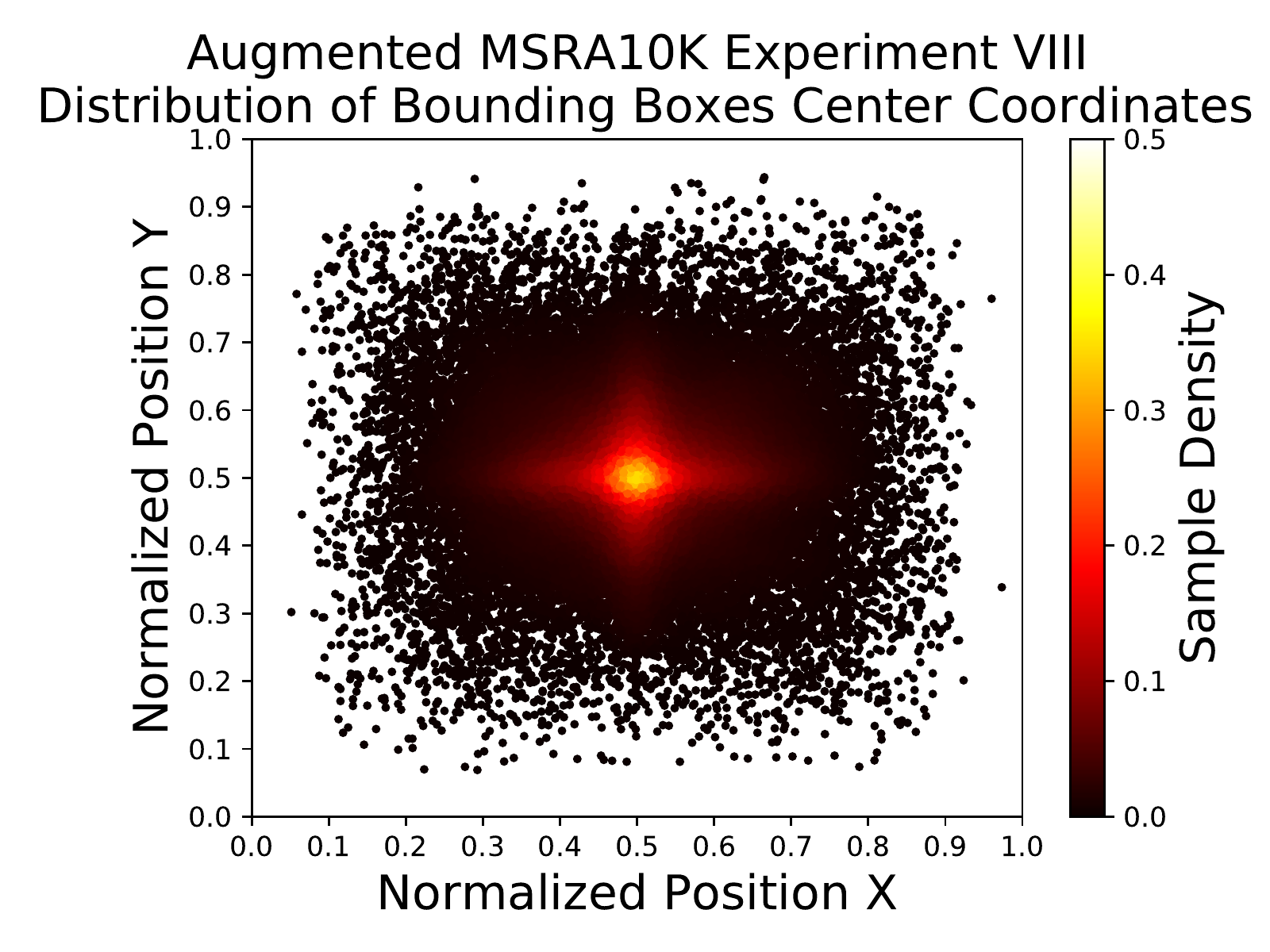}
\label{subfig:augpos}}
\hfil
\subfloat[]{\includegraphics[width=0.24\textwidth]{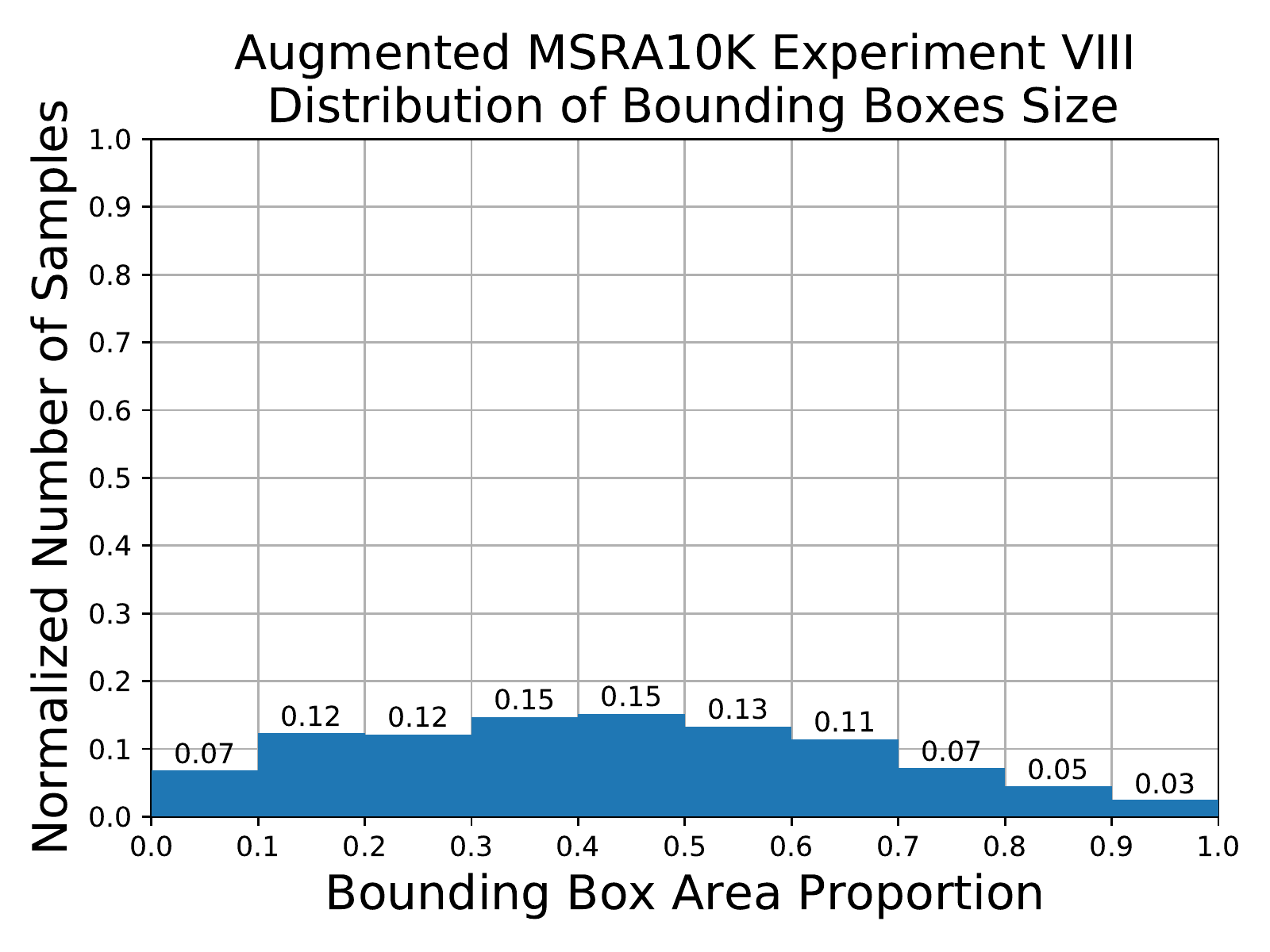}
\label{subfig:augsize}}

\caption{Position and size distributions per dataset. For position, the normalized bounding box center coordinates and heat colormap for sample position density, are presented as a scatter plot. For size, the bounding box area divided by image area is displayed in a 10-bin histogram. Figs~\ref{subfig:msra10kpos},~\ref{subfig:msra10ksize} correspond to the MSRA10K, the dataset used for training, and  Figs~\ref{subfig:augpos},~\ref{subfig:augsize} correspond to the augmented MSRA10K as defined by experiment VIII, see Section~\ref{sec:exp} for further details.}

\label{fig:dist}
\end{figure*}

\subsection{Size and Position Distribution}
\label{size_position}

The position and size distribution of the objects in the datasets were evaluated as follows. For each image, is generated the bounding box containing all objects inside the image. The position distribution analysis can be observed in a scatter plot where the center coordinates of the bounding box (normalized) are used, with a heat colormap for sample density, as shown in Fig.~\ref{subfig:msra10kpos}. The size distribution analysis can be observed in a ten bin histogram, where the area of the bounding box is divided by the area of the image to obtain the proportion of the bounding box over the entire image, as shown in Fig.~\ref{subfig:msra10ksize}. 

Since the MSRA10K dataset is the training dataset, the goal is to generate new samples with a high variation of position and size, approximating the distribution of the dataset to a uniform distribution. Since in this dataset there is a concentration of samples in the $[0.2,0.3), [0.3,0.4), [0.4,0.5)$ intervals, new samples don't need to be produced in those scales, neither in the $[0.9,1.0)$ interval, since an object in this size range, is not as useful for training, because it almost completely occlude the background. Since there is already a concentration on the center, for the position, the goal is to approximate the new samples to the margins, causing a dispersion on the samples.

The training dataset is composed of $k$ images. The index $i$ represents the $i$th image inside the training dataset. Each image $i$ has an object $o$, a background $b$, and a bounding box $\phi$ defined by two corner points $v=(x_{\min},y_{\min})=(x,y)$ and $v^\prime=(x_{\max},y_{\max})=(x^{\prime},y^{\prime})$. The bounding box has a width $w_\phi=x^\prime-x$, a height $h_\phi=y^\prime-y$, and an area $a_\phi=w_\phi*h_\phi$. The background has a width $w_b$, a height $h_b$, and an area $a_b=w_b*h_b$.

Aiming to resize the object $o$, we multiply its width and height to a scale factor $s$, if the resized object $o^\prime$ can fit $b$ then s is defined as in  Equation~\ref{eq:s} otherwise as in Equation~\ref{eq:s2}, generating a new object $o^{\prime}$ with $w_o^{\prime}=w_o*s$ 
and $h_o^{\prime}=h_o*s$. 

\begin{equation}
s = \sqrt{ \frac{p \times a_b}{a_\phi} }
\label{eq:s}
\end{equation}

If the resized object $o^\prime$ fits $b$, $s$ is as defined by Equation~\ref{eq:s}, where $p$ (Equation~\ref{eq:p}) is a scale factor to increase the number of objects with that given scale. $\beta$ defined at Equation~\ref{eq:beta} is a vector of scales, defined based on the lack of objects in those scales in the original dataset, as pointed out previously, and $\gamma$ is a vector of random values in the interval [0.975, 1.025] following a uniform distribution. The values in $\gamma$ are noises to generate objects close to the desired size, but not exactly as defined in $\beta$, increasing the diversity of object sizes.

\begin{equation}
    p =  \beta_{(i \mod 5)} \times \gamma_i
    \label{eq:p}
\end{equation}

\begin{equation}
    \beta = \begin{bmatrix} 0.05 & 0.15 & 0.65 & 0.75 & 0.85  \end{bmatrix}^T
    \label{eq:beta}
\end{equation}

\begin{equation}
   s = \min \left( \frac{w_b}{x^{\prime}-x},\frac{h_b}{y^{\prime}-y} \right) 
   \label{eq:s2}
\end{equation}

Otherwise, if the resized object $o^\prime$ cannot fit $b$, there are two cases, first case: both $\phi$ and $b$ have the same orientation, landscape or portrait, then s is defined as in Equation~\ref{eq:s2}, i.e., $s$ receives the highest possible value to resize the object without deforming it or overflowing the background. The second case, $\phi$, and $b$ have a different orientation, then $o$ is rotated by $-90^{\circ}$ degrees or $90^{\circ}$ degrees, the sign of the value is chosen based on a random binary value, then the same resize as the first case occurs.  

To disperse the objects center position in the dataset a translation is performed on $o$, in such a way that $\phi^\prime$ still in the boundaries of $b$, if there is free space in both directions for each axis. A random binary value decides the direction, the remaining space between $\phi^\prime$ and the boundary is then multiplied by $\theta_{tx}$ and $\theta_{tx}$ which are vectors of random values in the interval [0.0, 1.0] following a uniform distribution. An example of the effect of doing those operations is displayed in Fig.~\ref{fig:fullexample}.
\section{Experiments}\label{sec:exp}

In order to evaluate the network models, we utilize nine datasets of salient objects
widely used in the \gls*{sod} literature: DUT-OMRON~\cite{ruan:dutomron}, ECSSD~\cite{yan:ecssd},
HKU-IS~\cite{li:visual_saliency}, ICOSEG~\cite{icoseg}, MSRA10K~\cite{msra10k},
PASCAL-S~\cite{li:pascals}, SED1~\cite{borji:benchmark}, SED2~\cite{borji:benchmark}, and
THUR~\cite{ming:thur15k}. 

The datasets are composed by images containing salient objects with different biases (\emph{e.g.},
number of salient objects, image clutter, center-bias) and the referent ground truth mask with the
expected segmentation of the salient objects. The ground truth masks are binary images with the
salient regions in white (value 255) and the background with no salient regions in black (value 0).

\subsection{Evaluation Metrics}
The architecture models trained in the \gls*{sod} problem are evaluated and compared through four 
metrics widely used in the \gls*{sod} literature: Precision, Recall, F-measure also known as F-score, and \glsdisp*{mae}{Mean Absolute Error (MAE)}. The $\beta$ in the F-measure formula changes the Precision and Recall importance. In the \gls*{sod} literature, 
$\beta^{2}$ receives the value $0.3$~\cite{borji:benchmark,achanta:frequency_tuned}
to increase the Precision importance.

\subsection{Experiments}
\label{experiments}
We present our findings on four distinct neural networks. Three \glspl*{fcn} with \gls*{resnet}-50, \gls*{resnet}-101 and VGG-16 backbones, implemented on KittiSeg framework~\cite{teichmann:kittiseg} and the PoolNet~\cite{Liu2019PoolSal} with ResNet-50 backbone. We also compare our approach to other augmentation techniques proposed in the literature, such as horizontal flipping, rescale, rotation and random cropping.

The \glspl*{fcn} with \glspl*{resnet} 50 and 101 backbones were trained with 40,000 and 60,000
iterations respectively, receiving an initial learning rate of $10^{-6}$. We also utilized an exponential learning rate
decay, similar with the proposed in the literature~\cite{learning_rate_decay}, and the initial learning rate is divided
by 10 at the 50,000th iteration. The \gls*{fcn} with VGG-16 backbone was trained with 100,000 iterations,
receiving an initial learning rate of $10^{-5}$ which is divided by 10 at the 50,000th iteration,
generating a learning rate which is divided by 10 at the 75,000th. 

All the PoolNet experiments are
performed using a weight decay of $5\times10^{-4}$ and an
initial learning rate of $5\times10^{-5}$  which  is  divided  by  10  after  15  epochs (with an epoch
been $n$ iterations, with $n$ the number of images on the training dataset). Thus, each PoolNet experiment were trained for 24 epochs in total, both the baseline and the data-augmented ones were
trained with the same number of epochs. Joint Training with Edge Detection was not performed.

\scriptsize
\begin{table*}
\centering
\caption{
Comparison between baseline and data augmented results. Best F-score, Precision, and MAE per dataset are highlighted in \textbf{\textcolor{blue}{blue}}, \textbf{\textcolor{red}{red}}, and \textbf{\textcolor{orange}{orange}} respectively. DUT-O* is an abbreviation of DUT-OMRON.
}
\begin{tabularx}{\textwidth}{c *{10}{Y}}

\toprule
\multicolumn{1}{c}{\textbf{Experiment}} &
\multicolumn{1}{c}{\textbf{Metric}} &
\multicolumn{1}{c}{\textbf{DUT-O*}} &
\multicolumn{1}{c}{\textbf{ECSSD}} &
\multicolumn{1}{c}{\textbf{HKU-IS}} &
\multicolumn{1}{c}{\textbf{ICOSEG}} &
\multicolumn{1}{c}{\textbf{PASCAL-S}} &
\multicolumn{1}{c}{\textbf{SED1}} &
\multicolumn{1}{c}{\textbf{SED2}} &
\multicolumn{1}{c}{\textbf{THUR}} 
\\ \midrule
\makecell{Baseline} & 
\makecell{F-score \\ Precision \\ Recall \\ \gls*{mae}} & 
\makecell{0.499 \\ 0.681 \\ 0.441 \\ 0.106} &
\makecell{0.790 \\ 0.887 \\ 0.697 \\ 0.092} &
\makecell{0.761 \\ 0.870 \\ 0.665 \\ 0.073} &
\makecell{0.748 \\ 0.826 \\ 0.692 \\ 0.087} &
\makecell{0.714 \\ 0.823 \\ 0.646 \\ 0.104 } &
\makecell{0.641 \\ 0.890 \\ 0.497 \\ 0.129} &
\makecell{0.408 \\ 0.848 \\ 0.264 \\ 0.148} &
\makecell{0.617 \\ 0.691 \\ 0.649 \\ 0.099} 
\\ \addlinespace[0.5mm]
\makecell{(I) H-Flip\\\cite{Liu2019PoolSal,Bianco2017RC,Aytekin2018,GuoSymmetry2018,Huang2019,laroca2019}} & 
\makecell{F-score \\ Precision \\ Recall \\ \gls*{mae}} &
\makecell{0.499 \\ 0.661 \\ 0.444 \\ 0.102} &
\makecell{0.798 \\ 0.875 \\ 0.729 \\ \textbf{\textcolor{orange}{0.085}}} &
\makecell{0.762 \\ 0.854 \\ 0.691 \\ 0.071} &
\makecell{0.760 \\ 0.821 \\ 0.719 \\ 0.081} &
\makecell{0.717 \\ 0.808 \\ 0.672 \\ \textbf{\textcolor{orange}{0.100}}} &
\makecell{0.664 \\ 0.885 \\ 0.542 \\ \textbf{\textcolor{orange}{0.120}}} &
\makecell{0.386 \\ 0.780 \\ 0.258 \\ 0.148} &
\makecell{0.610 \\ 0.670 \\ 0.660 \\ 0.101} 
\\ \addlinespace[0.5mm]
\makecell{(II) Rotate\\\cite{GuoSymmetry2018,laroca2019}} & 
\makecell{F-score \\ Precision \\ Recall \\ \gls*{mae}} &
\makecell{0.513 \\ 0.689 \\ 0.461 \\ 0.105} &
\makecell{0.799 \\ 0.882 \\ 0.723 \\ 0.090} &
\makecell{0.789 \\ 0.858 \\ 0.733 \\ 0.066} &
\makecell{0.745 \\ 0.804 \\ 0.746 \\ 0.090} &
\makecell{0.726 \\ 0.825 \\ 0.669 \\ 0.101} &
\makecell{0.645 \\ 0.888 \\ 0.521 \\ 0.131} &
\makecell{0.481 \\ 0.860 \\ 0.336 \\ 0.138} &
\makecell{0.651 \\ 0.690 \\ 0.722 \\ 0.095} 
\\ \addlinespace[0.5mm]
\makecell{(III) Resize\\\cite{Perazzi2017CVPR,laroca2019}} & 
\makecell{F-score \\ Precision \\ Recall \\ \gls*{mae}} &
\makecell{0.504 \\ 0.689 \\ 0.437 \\ 0.102} &
\makecell{0.799 \\ 0.887 \\ 0.717 \\ 0.087} &
\makecell{0.770 \\ 0.867 \\ 0.685 \\ 0.069} &
\makecell{\textbf{\textcolor{blue}{0.765}} \\ \textbf{\textcolor{red}{0.840}} \\ 0.704 \\ \textbf{\textcolor{orange}{0.081}}} &
\makecell{0.721 \\ 0.825 \\ 0.659 \\ 0.102} &
\makecell{0.644 \\ 0.888 \\ 0.514 \\ 0.125} &
\makecell{0.411 \\ 0.875 \\ 0.266 \\ 0.143} &
\makecell{0.611 \\ 0.694 \\ 0.641 \\ 0.100} 
\\ \addlinespace[0.5mm]
\makecell{(IV) Crop\\\cite{Bianco2017RC,Huang2019,laroca2019}} & 
\makecell{F-score \\ Precision \\ Recall \\ \gls*{mae}} &
\makecell{0.523 \\ 0.687 \\ 0.473 \\ 0.104} &
\makecell{0.800 \\ 0.890 \\ 0.710 \\ 0.089} &
\makecell{0.779 \\ 0.873 \\ 0.691 \\ 0.069} &
\makecell{0.739 \\ 0.826 \\ 0.667 \\ 0.087} &
\makecell{0.726 \\ 0.830 \\ 0.656 \\ 0.102} &
\makecell{0.641 \\ 0.870 \\ 0.505 \\ 0.129} &
\makecell{0.409 \\ 0.880 \\ 0.271 \\ 0.144} &
\makecell{0.625 \\ 0.692 \\ 0.659 \\ 0.097} 
\\ \addlinespace[0.5mm]
\makecell{(V) ANDA $\leq0.1$} & 
\makecell{F-score \\ Precision \\ Recall \\ \gls*{mae}} &
\makecell{0.517 \\ \textbf{\textcolor{red}{0.713}} \\ 0.437 \\ 0.104} &
\makecell{0.798 \\ 0.899 \\ 0.695 \\ 0.096} &
\makecell{0.790 \\ \textbf{\textcolor{red}{0.887}} \\ 0.686 \\ 0.072} &
\makecell{0.753 \\ 0.832 \\ 0.692 \\ 0.099} &
\makecell{0.715 \\ 0.848 \\ 0.617 \\ 0.109} &
\makecell{0.622 \\ 0.916 \\ 0.483 \\ 0.141} &
\makecell{0.480 \\ 0.865 \\ 0.338 \\ 0.142} &
\makecell{0.646 \\ \textbf{\textcolor{red}{0.709}} \\ 0.681 \\ 0.096}
\\ \addlinespace[0.5mm]
\makecell{(VI) ANDA $\geq0.25$} & 
\makecell{F-score \\ Precision \\ Recall \\ \gls*{mae}} &
\makecell{0.530 \\ 0.702 \\ 0.469 \\ 0.101} &
\makecell{0.787 \\ 0.899 \\ 0.666 \\ 0.098} &
\makecell{0.768 \\ 0.885 \\ 0.650 \\ 0.076} &
\makecell{0.755 \\ 0.836 \\ 0.698 \\ 0.098} &
\makecell{0.707 \\ \textbf{\textcolor{red}{0.849}} \\ 0.599 \\ 0.112} &
\makecell{0.651 \\ 0.920 \\ 0.485 \\ 0.135} &
\makecell{0.478 \\ \textbf{\textcolor{red}{0.907}} \\ 0.309 \\ 0.144} &
\makecell{0.624 \\ 0.702 \\ 0.627 \\ 0.099}
\\ \addlinespace[0.5mm]
\makecell{(VII) ANDA \\ $\left \lfloor{ \frac{k}{2} }\right \rfloor$th neighboor} & 
\makecell{F-score \\ Precision \\ Recall \\ \gls*{mae}} &
\makecell{0.530 \\ 0.707 \\ 0.462 \\ \textbf{\textcolor{orange}{0.100}}} &
\makecell{0.799 \\ \textbf{\textcolor{red}{0.901}} \\ 0.697 \\ 0.093} &
\makecell{0.788 \\ 0.881 \\ 0.697 \\ 0.070} &
\makecell{0.755 \\ 0.823 \\ 0.723 \\ 0.093} &
\makecell{0.717 \\ 0.844 \\ 0.628 \\ 0.107} &
\makecell{0.683 \\ \textbf{\textcolor{red}{0.924}} \\ 0.545 \\ 0.130} &
\makecell{0.473 \\ 0.877 \\ 0.337 \\ 0.142} &
\makecell{0.646 \\ 0.706 \\ 0.682 \\ 0.095}
\\ \addlinespace[0.5mm]
\makecell{(VIII)} & 
\makecell{F-score \\ Precision \\ Recall \\ \gls*{mae}} &
\makecell{\textbf{\textcolor{blue}{0.549}} \\ 0.700 \\ 0.496 \\ 0.103} &
\makecell{\textbf{\textcolor{blue}{0.813}} \\ 0.885 \\ 0.734 \\ 0.089} &
\makecell{\textbf{\textcolor{blue}{0.803}} \\ 0.867 \\ 0.739 \\ \textbf{\textcolor{orange}{0.066}}} &
\makecell{0.758 \\ 0.811 \\ 0.739 \\ 0.092} &
\makecell{\textbf{\textcolor{blue}{0.730}} \\ 0.838 \\ 0.658 \\ 0.105} &
\makecell{\textbf{\textcolor{blue}{0.694}} \\ 0.917 \\ 0.573 \\ 0.124} &
\makecell{\textbf{\textcolor{blue}{0.549}} \\ 0.885 \\ 0.382 \\ \textbf{\textcolor{orange}{0.133}}} &
\makecell{\textbf{\textcolor{blue}{0.659}} \\ 0.688 \\ 0.737 \\ \textbf{\textcolor{orange}{0.095}}} 
\\ \bottomrule

\end{tabularx}
\label{tab:testComp}
\end{table*}
\normalsize

In the first step, all networks were trained in the MSRA10K dataset (baseline) without any alteration. Then, we generated eight new training sets with augmentation applied in the MSRA10K images:

\begin{enumerate}[(I)]
\item MSRA10K + horizontal flip (20,000 images in total).

\item MSRA10K + uniformly distributed random rotations between $[-30^\circ,30^\circ]$ (20,000).

\item MSRA10K + uniformly distributed random resizes between $[0.9,1.1]$ (20,000).

\item MSRA10K + uniformly distributed random crop preserving the salient object (20,000).

\item MSRA10K + ANDA technique limiting the distance between the object and the background to less than $0.1$, considering the cosine similarity as previously discussed in Section~\ref{knn} (20,000).

\item MSRA10K + ANDA technique limiting the distance between the object and the background to greater than $0.25$ (20,000).

\item MSRA10K + ANDA technique taking the background with
$\left \lfloor{ \frac{k}{2} }\right \rfloor$th nearest neighbor (20,000).

\item In this experiment, let $\Omega$ be the union set of the augmentation IV and VII, and $\omega$ a uniformly distributed random selection of 15,000 images from $\Omega$. In $\omega$ was performed horizontal flip, a uniformly distributed random rescale of $[0.9,1.1]$, and uniformly distributed random rotation of $[-30^\circ,30^\circ]$. The Experiment VIII is composed of MSRA10K + $\Omega$ + $\omega$ (45,000 images in total).

\end{enumerate}

\scriptsize
\begin{table*}
\centering
\caption{
Comparison between baseline and data augmented results. Best F-score and MAE per network are highlighted in \textbf{bold text}. DUT-O* is an abbreviation of DUT-OMRON.
}
\begin{tabularx}{\textwidth}{c *{11}{Y}}

\toprule
\multicolumn{1}{c}{\textbf{Network}} &
\multicolumn{1}{c}{\textbf{Experiment}} &
\multicolumn{1}{c}{\textbf{Metric}} &
\multicolumn{1}{c}{\textbf{DUT-O*}} &
\multicolumn{1}{c}{\textbf{ECSSD}} &
\multicolumn{1}{c}{\textbf{HKU-IS}} &
\multicolumn{1}{c}{\textbf{ICOSEG}} &
\multicolumn{1}{c}{\textbf{PASCAL-S}} &
\multicolumn{1}{c}{\textbf{SED1}} &
\multicolumn{1}{c}{\textbf{SED2}} &
\multicolumn{1}{c}{\textbf{THUR}} 
\\ \midrule

\multirow{6}{*}{
\makecell{\gls*{fcn} \\ \gls*{resnet}-101}} & 
\makecell{Baseline} & 
\makecell{F-score \\ \gls*{mae}} &
\makecell{0.442 \\ 0.107} &
\makecell{0.766 \\ \textbf{0.094}} &
\makecell{0.735 \\ 0.075} &
\makecell{0.682 \\ 0.108} &
\makecell{\textbf{0.691} \\ \textbf{0.107}} &
\makecell{\textbf{0.538} \\ \textbf{0.152}} &
\makecell{0.315 \\ 0.186} &
\makecell{0.576 \\ 0.102}
\\ \addlinespace[0.5mm]
\makecell{} & 
\makecell{VII}&
\makecell{F-score \\ \gls*{mae}} &
\makecell{0.440 \\ 0.104} &
\makecell{0.761 \\ 0.103} &
\makecell{0.731 \\ 0.082} &
\makecell{0.643 \\ 0.124} &
\makecell{0.657 \\ 0.123} &
\makecell{0.439 \\ 0.170} &
\makecell{0.314 \\ 0.164} &
\makecell{0.586 \\ 0.097} 
\\ \addlinespace[0.5mm]
\makecell{} & 
\makecell{VIII}&
\makecell{F-score \\ \gls*{mae}} &
\makecell{\textbf{0.469} \\ \textbf{0.101}} &
\makecell{\textbf{0.784} \\ 0.096} &
\makecell{\textbf{0.771} \\ \textbf{0.072}} &
\makecell{\textbf{0.696} \\ 0.112} &
\makecell{0.685 \\ 0.114} &
\makecell{0.490 \\ 0.161} &
\makecell{\textbf{0.345} \\ \textbf{0.160}} &
\makecell{\textbf{0.620} \\ \textbf{0.094}}
\\ \midrule
\multirow{6}{*}{
\makecell{VGG-16}} &
\makecell{Baseline} & 
\makecell{F-score \\ \gls*{mae}} &
\makecell{\textbf{0.695} \\ \textbf{0.067}} &
\makecell{\textbf{0.863} \\ \textbf{0.081}} &
\makecell{0.840 \\ \textbf{0.064}} &
\makecell{\textbf{0.806} \\ 0.094} &
\makecell{\textbf{0.797} \\ \textbf{0.092}} &
\makecell{\textbf{0.858} \\ \textbf{0.079}} &
\makecell{0.590 \\ 0.134} &
\makecell{\textbf{0.722} \\ \textbf{0.074}} 
\\ \addlinespace[0.5mm]
\makecell{} &
\makecell{VII}&
\makecell{F-score \\ \gls*{mae}} &
\makecell{0.664 \\ 0.078} &
\makecell{0.850 \\ 0.084} &
\makecell{0.839 \\ 0.066} &
\makecell{0.793 \\ 0.096} &
\makecell{0.777 \\ 0.099} &
\makecell{0.823 \\ 0.086} &
\makecell{0.616 \\ 0.134} &
\makecell{0.710 \\ 0.078} 
\\ \addlinespace[0.5mm]
\makecell{} &
\makecell{VIII}&
\makecell{F-score \\ \gls*{mae}} &
\makecell{0.687 \\ 0.070} &
\makecell{0.857 \\ 0.081} &
\makecell{\textbf{0.844} \\ 0.065} &
\makecell{0.799 \\ \textbf{0.092}} &
\makecell{0.788 \\ 0.095} &
\makecell{0.848 \\ 0.080} &
\makecell{\textbf{0.647} \\ \textbf{0.125}} &
\makecell{0.719 \\ 0.075}
\\ \midrule
\multirow{6}{*}{
\makecell{PoolNet \\ \gls*{resnet}-50}} & 
\makecell{Baseline} & 
\makecell{F-score \\ \gls*{mae}} &
\makecell{\textbf{0.737} \\ \textbf{0.060}} &
\makecell{0.903 \\  0.049} &
\makecell{\textbf{0.893} \\ \textbf{0.036}} &
\makecell{\textbf{0.844} \\  0.071} &
\makecell{\textbf{0.844} \\ \textbf{0.068}} &
\makecell{0.906 \\ 0.047} &
\makecell{0.815 \\ 0.085} &
\makecell{\textbf{0.726} \\ \textbf{0.073}} 
\\ \addlinespace[0.5mm]
\makecell{} & 
\makecell{VII}&
\makecell{F-score \\ \gls*{mae}} &
\makecell{0.735 \\ 0.066} &
\makecell{0.901 \\ 0.048} &
\makecell{0.889 \\ 0.037} &
\makecell{0.831 \\ 0.071} &
\makecell{0.839 \\ 0.070} &
\makecell{0.912 \\ 0.045} &
\makecell{0.814 \\ 0.077} &
\makecell{0.721 \\ 0.075} 
\\ \addlinespace[0.5mm]
\makecell{} & 
\makecell{VIII} & 
\makecell{F-score \\ \gls*{mae}} &
\makecell{0.725 \\ 0.072} &
\makecell{\textbf{0.904} \\ \textbf{0.046}} &
\makecell{0.884 \\ 0.038} &
\makecell{0.829 \\ \textbf{0.070}} &
\makecell{0.837 \\ {0.070}} &
\makecell{\textbf{0.915} \\ \textbf{0.041}} &
\makecell{\textbf{0.845} \\ \textbf{0.068}} &
\makecell{0.717 \\ 0.078}  
\\ \bottomrule

\end{tabularx}
\label{tab:tests}
\end{table*}
\normalsize

\section{DISCUSSION}

To ensure a fair comparison between our proposed approach and the commonly used techniques on the literature, we isolate each technique, defined the parameters, further details in Section~\ref{sec:exp}, and compared one another on the same~network. 

In Table~\ref{tab:testComp} we present the results achieved with the baseline (MSRA10K), and in the eight augmentations performed
in our work in the \gls*{fcn} with \gls*{resnet}-50 backbone. The proposed augmentation ANDA achieved the highest F-measure in the
DUT-OMRON, SED1, SED2, and achieved the highest Precision in all datasets, with exception of the ICOSEG. In the last augmentation
(VIII), where we combined the proposed ANDA with other augmentation techniques, we achieved the highest F-measures in seven of eight
datasets and achieved the smallest \gls*{mae} in three of eight datasets. Experiment VII outperformed V and VI in all datasets when evaluating the MAE encouraging our previous assumption of Section~\ref{knn}.

In Table~\ref{tab:tests} we present a comparison of three networks, a \gls*{fcn} with \gls*{resnet}-101 backbone, a \gls*{fcn} with
VGG-16 backbone and PoolNet with \gls*{resnet}-50 backbone. The three networks were trained with three training sets: the MSRA10K
(baseline), our proposed ANDA (augmentation VII), and our proposed ANDA with other augmentation techniques (VIII). With the
\gls*{fcn} with \gls*{resnet}-101 backbone, the augmentation VIII achieved the highest F-measure in six of eight datasets, and achieved
the smallest \gls*{mae} in four datasets. Our method was not so effective with the \gls*{fcn} with VGG-16 backbone,  and improved the
F-measure only in two datasets, when compared with the same network trained with the baseline. Finally, with the PoolNet, the 
augmentation VIII achieved higher F-measure in three datasets, and achieved better \gls*{mae} in four datasets.

\section{Conclusion}

In this work, we proposed a novel data augmentation technique (ANDA) for the Visual Saliency problem
and performed experiments on eight different training sets, where  the combination of diverse
data augmentation techniques widely utilized in the \gls*{sod} literature and our technique are compared. For each network, cross-dataset tests were performed in eight different publicly available datasets. Each experiment repeated those tests varying only the training set. The experiments have shown that
the ANDA technique when combined with other conventional approaches, such as random cropping, horizontal
flipping, rotation, and re-scale, provides improvements even in the PoolNet, a very recent state of the
art network. Further exploration of this technique on other networks and contexts was left for
future works.




\section*{ACKNOWLEDGMENT}
The authors would like to thank the Coordination for the Improvement of Higher Education Personnel
(CAPES) for the Masters scholarship. We gratefully acknowledge the founders of the publicly
available datasets and the support of NVIDIA Corporation with the donation of the GPUs used for this research.


\balance
\bibliographystyle{IEEEtran}
\bibliography{root}

\end{document}